\documentclass[11pt]{article}

\usepackage[preprint]{acl}

\usepackage{times}
\usepackage{latexsym}

\usepackage[T1]{fontenc}
\usepackage[utf8]{inputenc}

\usepackage{microtype}
\usepackage{inconsolata}

\usepackage{graphicx}

\usepackage{booktabs}
\usepackage{array}
\usepackage{multirow}

\usepackage{amsmath}
\usepackage{amssymb}
\usepackage{amsfonts}

\usepackage{algorithm}
\usepackage{algpseudocode}

\usepackage{fancyvrb}
\fvset{fontsize=\footnotesize,xleftmargin=0pt}

\title{Denoising Iterative Self-Correction:\\Structured Verification Loops for Reliable LLM Reasoning}

\author{
Shen Yin \and David Ken \and Joel Stremmel \\
Thomson Reuters Labs
}
  
\begin{document}
\maketitle

\begin{abstract}
Large language models produce fluent but often incorrect multi-step reasoning, and naive correction methods risk degrading already-correct answers. We introduce \textbf{Denoising Iterative Self-Correction (DISC)}, a test-time procedure that treats verification question outputs as noisy measurements of where a solution may be corrupted. Using these signals, DISC progressively reduces errors across multiple verify-judge-correct passes, analogous to traditional iterative denoising. A binary judgment gate controls correction precision by blocking rewrites that would damage already-correct answers while the verifier and corrector together repair errors. We evaluate this trade-off using two paired diagnostics: an improvement-to-degradation ratio (precision) and a repair rate (recall). Across three benchmarks (BIG-Bench Mistake, HotpotQA, GPQA Diamond) and four models, DISC dominates Chain-of-Verification and Self-Refine on the precision-recall trade-off, reaching 81.6\% accuracy with $13\times$ more improvements per degradation than Chain-of-Verification and $5\times$ more than Self-Refine on BIG-Bench Mistake (Sonnet~4.5). On GPQA Diamond, we identify a capability floor below which judges acknowledge contradictions in evidence but cannot translate that recognition into a correction. We further show that cross-model role allocation---assigning verification and judgment to a model different from the generator---mitigates self-confirmation bias.
\end{abstract}

\section{Introduction}
\label{sec:introduction}

Large language models increasingly operate in settings that demand correctness, such as decision support, analytics, deep research, trend analysis, and professional drafting. Yet LLM outputs frequently contain errors, particularly in tasks requiring multi-step transformations or multi-hop evidence aggregation \citep{tyen-etal-2024-llms,huang2024large}. A single arithmetic mistake in step three of a ten-step derivation can invalidate the entire reasoning chain; a missed qualifier in multi-hop question answering can lead to a confidently stated wrong answer.

Several broad paradigms address this reliability gap. \textbf{Prompting-based} approaches such as chain-of-thought \citep{wei2022chain} elicit step-by-step reasoning that improves accuracy but does not necessarily detect and correct mistakes. \textbf{Sampling and selection} methods, such as self-consistency \citep{wang2023self}, generate multiple independent solutions and select among them by majority vote. \textbf{Test-time correction} methods attempt to improve a single initial answer through additional verification and revision passes. \textbf{Training-based} methods modify the model itself, through reinforcement learning \citep{kumar2024scoring}, reward model training \citep{ouyang2022training}, or alignment objectives \citep{rafailov2023direct}.

This paper focuses on test-time correction. A critical but underreported requirement for any correction procedure is that it should not frequently degrade correct answers. A method that repairs 10 errors but introduces 8 new ones is unreliable , even when its net accuracy gain is positive. We therefore frame correction evaluation in standard classification terms: each correction attempt is a decision about whether to change an answer, and a method's quality is jointly determined by its precision (the ratio of helpful corrections to harmful ones) and its recall (what fraction of fixable errors it actually fixes). We operationalize these as the \textbf{improvement-to-degradation ratio} ($I{:}D$) and the \textbf{repair rate}, the latter being the single-sample empirical form of the Critique Score of \citet{yang2025confidence}; we report them as paired quantities throughout, with formal definitions in Section~\ref{sec:metrics}. While prior work has surfaced degradation as a distinct concern \citep{huang2024large,yang2025confidence}, accuracy and net change remain the most common headline metrics in the self-correction literature \citep{madaan2023selfrefine,gou2024critic}.

Our work is motivated by a simple observation: verification is noisy. A single round of checking may miss subtle failure modes, over-focus on irrelevant details, or hallucinate discrepancies that do not exist. We borrow intuition from iterative denoising in image restoration and diffusion models \citep{ho2020denoising,song2019generative}, where each step removes a fraction of the noise while preserving signal that is consistent across passes. DISC treats verification outputs (questions, evidence, and judgments) as noisy measurements of where the current solution may be corrupted, and therefore repeats the verify-judge-correct procedure across multiple iterations to progressively reduce errors while preserving parts that remain consistent with the evidence. 

\begin{figure}[t]
  \centering
  \includegraphics[width=\columnwidth]{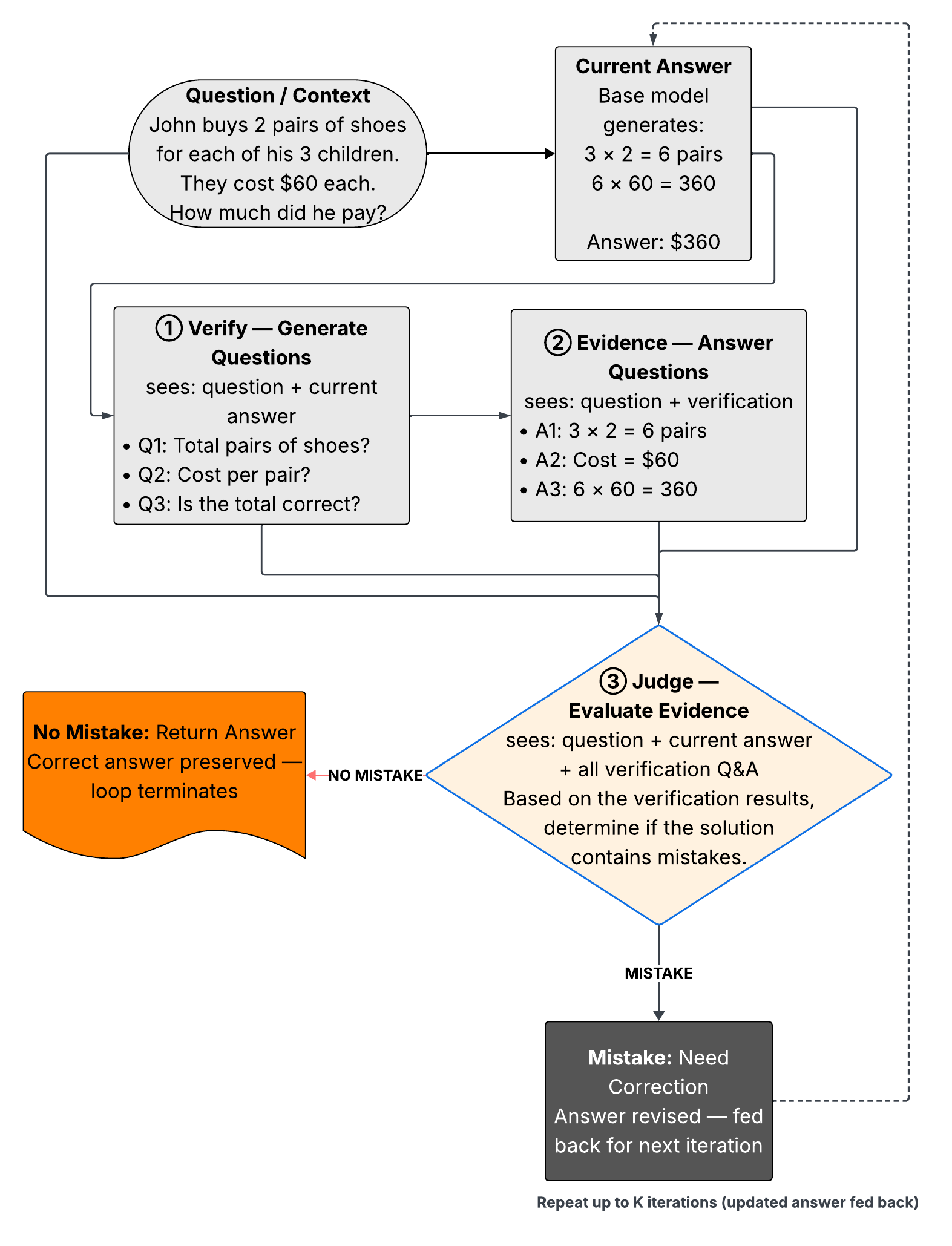}
  \caption{DISC pipeline. Given an initial answer $y_0$, the system repeats: (1)~generate verification questions targeting potential errors in $y_t$, (2)~answer questions to produce evidence, (3)~judge whether $y_t$ contains a mistake. If \textsc{No\_Mistake}, return $y_t$. If \textsc{Mistake}, (4)~correct $y_t$ using the evidence.}
  \label{fig:pipeline}
\end{figure}

\textbf{DISC overview.} DISC iterates a verify--judge--correct loop (Figure~\ref{fig:pipeline}; Section~\ref{sec:pipeline}) with an explicit judgment gate: when the judge finds no mistake, the loop terminates early and the answer is returned unchanged. Early termination is the common case (e.g., on GPQA Diamond cross-model at $K{=}3$, 79\% of examples reach a no-mistake verdict and exit before exhausting the iteration budget). We show in Section~\ref{sec:gate-ablation} that this gate functions as the system's precision mechanism.

\textbf{Contributions.} (1)~the DISC framework with an explicit judgment gate that serves as the system's precision mechanism, separating mistake detection from correction and enabling early stopping and principled preservation of correct answers; (2)~an evaluation spanning three benchmarks chosen for task diversity: algorithmic reasoning (BIG-Bench Mistake), multi-hop question answering (HotpotQA), and graduate-level science (GPQA Diamond); (3)~cross-model configurations identifying self-confirmation bias as the primary obstacle to same-model self-correction and cross-model role allocation as an effective remedy; and (4)~evidence of a minimum capability threshold below which models acknowledge contradictions in evidence but cannot translate that recognition into corrective action.

\section{Related Work}
\label{sec:related-work}

\textbf{Principal baselines.} We compare DISC against Chain-of-Verification (CoVe) \citep{dhuliawala2023chain}, which generates verification questions and answers to refine the initial response in a single pass, and Self-Refine \citep{madaan2023selfrefine}, in which a single LLM iteratively critiques and revises its output until a stopping criterion is met. Relative to CoVe, DISC adds a binary judgment gate before correction (avoiding unconditional revision) and iterates the verify-judge-correct process; relative to Self-Refine, it replaces holistic critique with factored verification questions answered separately. 

\textbf{Evaluation benchmarks.} BIG-Bench Mistake (BBM) \citep{tyen-etal-2024-llms} was constructed explicitly to test reasoning-error detection and correction: it contains pre-generated chain-of-thought traces with human-annotated first-mistake locations across five subtasks. We treat BBM as our primary correction benchmark and conduct our deepest analyses on it, including per-subtask results (Appendix~\ref{sec:appendix-bbm-pertask}) and the judgment gate ablation (Section~\ref{sec:gate-ablation}). We do not use BBM's accompanying backtracking method as a baseline because it feeds ground-truth mistake locations to the model, making it an oracle rather than a self-correction baseline (see Section~\ref{sec:baselines}). We complement BBM with HotpotQA and GPQA Diamond for multi-hop QA and graduate-level science.

\textbf{Critical perspectives.} \citet{huang2024large} and \citet{kamoi2024when} argue that intrinsic self-correction with same-model feedback rarely improves reasoning, locating the bottleneck in feedback generation. Our same-model and cross-model experiments test this position directly (Sections~\ref{sec:main-results}, \ref{sec:gpqa-fiveconfig}, \ref{sec:capability-floor}).

\textbf{Verifier bottleneck.} \citet{zhang2024small} propose that self-correction is ``largely bottlenecked by the verifier rather than the refiner,'' and that small language models benefit most from strong external verifiers. We investigate both claims directly through cross-model role allocation on GPQA (Section~\ref{sec:gpqa-fiveconfig} and Appendix~\ref{sec:appendix-confusion}) and through the gate ablation, which further isolates the verifier-corrector loop's contribution from the gate's (Section~\ref{sec:gate-ablation}).

\textbf{Training-based and tool-augmented correction.} SCoRe \citep{kumar2024scoring} trains models for self-correction via multi-turn online RL; CRITIC \citep{gou2024critic} grounds verification in external tools (code interpreters, search). Both are complementary to DISC, which operates purely at test time with model-generated verification.

\textbf{Confidence vs.\ critique.} \citet{yang2025confidence} decompose self-correction into ``confidence'' (preserving correct answers) and ``critique'' (repairing incorrect ones), and report two within-class rates: Confidence Level (CL) and Critique Score (CS). Our repair rate is the single-sample empirical form of CS, and our framing of correction quality as preservation-plus-repair follows their decomposition directly. Our additions are (i)~an improvement-to-degradation ratio (I:D) that combines preservation and repair into a single trade-off summary, and (ii)~the systematic use of I:D and repair rate as a precision-recall lens for comparing correction methods (Yang et al.\ use CL/CS to characterize individual models). DISC's judgment gate primarily affects I:D; its iteration loop primarily affects repair rate.

\section{Method: Denoising Iterative Self-Correction}

\subsection{Problem Setting}

Given an input $x$ (question, context, or task description) and a base language model $\mathcal{M}$, we produce an initial answer $y_0 \sim \mathcal{M}(x)$. The goal is to obtain a refined answer $y_K$ after up to $K$ iterations that maximizes task accuracy while minimizing degradation, the probability that a correct $y_0$ becomes an incorrect $y_K$.

\subsection{Pipeline}
\label{sec:pipeline}

DISC iterates the verify--judge--correct loop given in Algorithm~\ref{alg:disc}. Four roles, generator $\mathcal{M}_g$, verifier $\mathcal{M}_v$, judge $\mathcal{M}_j$, and corrector $\mathcal{M}_c$, are applied in sequence for up to $K$ iterations, with early termination whenever the judge returns \textsc{No\_Mistake}.

\begin{algorithm}[t]
\caption{Denoising Iterative Self-Correction (DISC).}
\label{alg:disc}
\small
\begin{algorithmic}[1]
\Require Input $x$; iteration budget $K$; role models $\mathcal{M}_g$ (generator), $\mathcal{M}_v$ (verifier), $\mathcal{M}_j$ (judge), $\mathcal{M}_c$ (corrector)
\Ensure Refined answer $y$
\State $y_0 \gets \mathcal{M}_g(x)$ \Comment{initial answer}
\For{$t = 0, 1, \ldots, K{-}1$}
    \State $Q_t \gets \mathcal{M}_v(x, y_t)$ \Comment{verification questions}
    \State $E_t \gets \mathcal{M}_v(x, Q_t)$ \Comment{answers to $Q_t$}
    \State $(d_t, r_t) \gets \mathcal{M}_j(x, y_t, Q_t, E_t)$ \Comment{decision $d_t \in \{\textsc{Mistake}, \textsc{No\_Mistake}\}$, rationale $r_t$}
    \If{$d_t = \textsc{No\_Mistake}$}
        \State \Return $y_t$ \Comment{early stop}
    \EndIf
    \State $y_{t+1} \gets \mathcal{M}_c(x, y_t, r_t)$ \Comment{evidence-conditioned rewrite}
\EndFor
\State \Return $y_K$
\end{algorithmic}
\end{algorithm}

The verifier produces a set of targeted questions probing specific claims or reasoning steps in $y_t$, then answers each independently. The judge requires a specific contradiction between $E_t$ and $y_t$ before flagging a mistake, prioritizing precision over recall to preserve already-correct answers; the accompanying rationale $r_t$ points to the offending evidence. The corrector treats the rationale as advisory, verifies that the flagged issue affects the final answer before rewriting, and leaves unflagged content unchanged. This evidence-conditioned design focuses revisions on identified errors rather than encouraging wholesale rewrites. Each iteration plays the role of one denoising step in the analogy, progressively reducing the remaining error.

\textbf{Iteration budget.} All main experiments use $K{=}3$. An ablation on GPQA Diamond (Section~\ref{sec:ablation-iter}) finds that $K{=}1 \to K{=}3$ produces large gains in net correction while $K{=}3$ onward plateaus; $K{=}3$ captures the bulk of available correction headroom at bounded per-example cost.

\textbf{Cross-model configurations.} The four roles need not share a backbone. Our experiments show that pairing a stronger verifier and judge with a weaker generator and corrector avoids self-confirmation bias while keeping generation costs low (Section~\ref{sec:capability-floor}).

\subsection{Prompt Design}

Prompts assign roles, provide inputs, and specify output formats, without few-shot examples or chain-of-thought scaffolding. Full templates appear in Appendix~\ref{sec:appendix-prompts}. Beyond Algorithm~\ref{alg:disc}, the only design choice is that the verifier produces a small number of targeted questions (typically 3--10) and answers each one independently of the current answer $y_t$.

\section{Experimental Setup}

\subsection{Benchmarks and Models}
\label{sec:benchmarks}

We evaluate DISC across three English-language benchmarks summarized in Table~\ref{tab:benchmarks}, spanning algorithmic reasoning, multi-hop question answering, and graduate-level science, using four models: Claude Sonnet 4.5, gpt-5.2, GPT-4o, and gpt-4.1-nano. BIG-Bench Mistake and HotpotQA use Sonnet 4.5 and GPT-4o; GPQA Diamond pairs gpt-4.1-nano and gpt-5.2. Built-in reasoning/thinking modes are disabled across all models, since DISC itself is the test-time correction mechanism under evaluation (Appendix~\ref{sec:appendix-model-settings}).

\begin{table*}[t]
\centering
\footnotesize
\begin{tabular}{llp{3.7cm}p{3.4cm}p{4.0cm}}
\toprule
\textbf{Benchmark} & \textbf{N} & \textbf{Domain / Subtask} & \textbf{Answer Format} & \textbf{Task} \\
\midrule
\multirow{5}{*}{BIG-Bench Mistake}
  & 300     & Tracking Shuffled Objects & Object name          & Track items through swaps \\
  & 300     & Logical Deduction         & Letter               & Order items by deduction \\
  & 300     & Multistep Arithmetic      & Integer              & Evaluate arithmetic expression \\
  & 300     & Word Sorting              & Space-separated list & Sort words alphabetically \\
  & 592     & Dyck Languages            & Bracket string       & Complete balanced delimiters \\
\midrule
GPQA Diamond & 198     & Graduate-level science (3 main domains) & Multiple choice (4-way) & Answer multiple-choice questions \\
HotpotQA     & 7{,}405 & Multi-hop QA over Wikipedia             & Free text               & Answer using supporting facts \\
\bottomrule
\end{tabular}
\caption{Benchmarks used in our evaluation. BIG-Bench Mistake totals 1{,}792 examples across five subtasks; per-subtask sizes are shown.\protect\footnotemark}
\label{tab:benchmarks}
\end{table*}
\footnotetext{Per-method evaluation sizes can be slightly lower than the canonical totals when individual examples are blocked by API content-policy filters. Per-method effective sizes are reported in the per-task tables in Appendix~\ref{sec:appendix-bbm-pertask}.}

\subsection{Baselines}
\label{sec:baselines}

Chain-of-Verification (CoVe) \citep{dhuliawala2023chain} and Self-Refine \citep{madaan2023selfrefine} are used as baselines and are described in Section~\ref{sec:related-work}; we additionally report the uncorrected initial answer as a reference. For CoVe we use the \emph{factored} variant, the best-performing CoVe variant on short-form QA tasks per \citet{dhuliawala2023chain}. For Self-Refine we use the prompt templates and stopping rule from the original implementation. DISC's prompts are rule-based to express the judgment gate's decision criteria, while CoVe's are few-shot following its original implementation; the judgment gate is an architectural choice (CoVe has no analog). The CoVe authors demonstrated stronger performance than the popular chain-of-thought \citep{wei2022chain} reasoning method, perhaps the most natural correction approach for LLMs; hence, we omit a chain-of-thought baseline.

\subsection{Metrics}
\label{sec:metrics}

We report \textbf{accuracy} (fraction of correct final answers) plus four correction-quality metrics computed against the initial answer: \textbf{improvement} ($I$, initially incorrect answers turned correct), \textbf{degradation} ($D$, initially correct answers turned incorrect), \textbf{net change} ($I-D$), and two diagnostics that operationalize the precision-recall framing. The \textbf{improvement-to-degradation ratio}, $I{:}D = I/D$, measures correction \emph{precision}: how many initially incorrect answers the method repairs for every initially correct answer it breaks. Values below~1 indicate net-negative correction. The \textbf{repair rate}, $I/N_{\text{wrong init}}$, measures correction \emph{recall}: of the examples whose initial answer was wrong (and were thus fixable), what fraction did the method actually fix. Under single-sample evaluation, the repair rate coincides with \citet{yang2025confidence}'s Critique Score. The two are independent in principle: a method can have high precision by attempting few corrections (high $I{:}D$, low repair) or high recall by attempting many (high repair, low $I{:}D$). Reporting them together surfaces this trade-off, which net accuracy alone obscures. For final accuracy, token-level F1, and repair rate we report 95\% bootstrap confidence intervals computed over 1{,}000 resamples on the per-example outcomes. Count-based metrics ($I$, $D$, $I{:}D$, Net) are reported as point estimates because they can be reconstructed from the underlying rates and example count. Ablation tables are reported as point estimates only.

\section{Results}
\label{sec:experiments}

Results are organized around three claims: DISC against baselines across our three primary benchmarks (Section~\ref{sec:main-results}), the role of model capability in verification (Section~\ref{sec:cross-model}), and DISC-internal ablations (Section~\ref{sec:ablations}).

\subsection{DISC vs.\ Baselines Across Benchmarks}
\label{sec:main-results}

Table~\ref{tab:main-results} reports DISC against CoVe and Self-Refine in same-model (one model fills every pipeline role) configurations across three primary benchmarks. 

\begingroup
\renewcommand{\topfraction}{0.95}
\renewcommand{\textfraction}{0.05}
\setlength{\textfloatsep}{8pt plus 2pt minus 2pt}
\begin{table*}[!t]
\centering
\footnotesize
\setlength{\tabcolsep}{3.5pt}
\renewcommand{\arraystretch}{1.1}
\begin{tabular}{lllcccccc}
\toprule
\textbf{Dataset} & \textbf{Model} & \textbf{Method} & \textbf{Init} & \textbf{Final} & \textbf{Imp.} & \textbf{Deg.} & \textbf{I:D} & \textbf{Repair} \\
\midrule
\multirow{6}{*}{\shortstack[l]{BIG-Bench\\Mistake\\\scriptsize$N{=}1{,}790$}}
  & Claude Sonnet 4.5 & DISC ($K{=}3$)  & 15.0 & \textbf{81.6} \tiny{[79.7, 83.4]} & 1{,}195 & 8   & \textbf{149:1}  & \textbf{81.3} \tiny{[79.3, 83.2]} \\
  &                   & CoVe            & 15.0 & 58.6 \tiny{[56.3, 60.9]}          & 851     & 75  & 11.3:1 & 58.0 \tiny{[55.6, 60.4]} \\
  &                   & Self-Refine     & 15.0 & 49.6 \tiny{[47.2, 51.7]}          & 633     & 21  & 30.1:1 & 42.0 \tiny{[39.4, 44.3]} \\
  & GPT-4o            & DISC ($K{=}3$)  & 15.0 & 49.9 \tiny{[47.6, 52.2]}          & 666     & 18  & 37:1   & 43.7 \tiny{[41.0, 46.3]} \\
  &                   & CoVe            & 15.0 & 62.0 \tiny{[59.9, 64.3]}          & 916     & 45  & 20.4:1 & 60.2 \tiny{[57.7, 62.8]} \\
  &                   & Self-Refine     & 15.0 & 27.6 \tiny{[25.7, 29.5]}          & 289     & 57  & 5.1:1  & 18.5 \tiny{[16.5, 20.5]} \\
\midrule
\multirow{6}{*}{\shortstack[l]{GPQA\\Diamond\\\scriptsize$N{=}198$}}
  & gpt-4.1-nano & DISC ($K{=}3$) & 54.0 & 54.0 \tiny{[47.5, 61.1]}          & 0  & 0  & ---    & 0.0 \tiny{[0.0, 0.0]}   \\
  &              & CoVe           & 54.0 & 53.0 \tiny{[46.0, 60.1]}          & 9  & 11 & 0.82:1 & 9.9 \tiny{[4.4, 16.5]} \\
  &              & Self-Refine    & 54.0 & 54.5 \tiny{[47.5, 61.1]}          & 3  & 2  & 1.5:1  & 3.3 \tiny{[0.0, 6.6]}   \\
  & gpt-5.2      & DISC ($K{=}3$) & 75.3 & \textbf{79.3} \tiny{[73.2, 84.8]} & 18 & 10 & \textbf{1.8:1}  & \textbf{36.7} \tiny{[22.4, 49.0]} \\
  &              & CoVe           & 75.3 & 69.7 \tiny{[63.6, 75.8]}          & 17 & 28 & 0.61:1 & 34.7 \tiny{[22.4, 49.0]} \\
  &              & Self-Refine    & 75.3 & 75.3 \tiny{[69.2, 81.3]}          & 13 & 13 & 1:1    & 26.5 \tiny{[14.3, 38.8]} \\
\midrule
\multirow{6}{*}{\shortstack[l]{HotpotQA\\\scriptsize$N{=}7{,}405$}}
  & Claude Sonnet 4.5 & DISC ($K{=}3$) & 65.1 & \textbf{80.4} \tiny{[79.7, 81.1]} & 1{,}487 & 109 & 13.6:1 & 36.8 \tiny{[35.4, 38.3]} \\
  &                   & CoVe           & 65.1 & 80.0 \tiny{[79.3, 80.8]}          & 1{,}687 & 224 & 7.5:1  & \textbf{41.8} \tiny{[40.3, 43.2]} \\
  &                   & Self-Refine    & 65.1 & 77.8 \tiny{[77.0, 78.5]}          & 1{,}085 & 28  & \textbf{38.8:1} & 26.9 \tiny{[25.5, 28.2]} \\
  & GPT-4o            & DISC ($K{=}3$) & 75.0 & 77.9 \tiny{[77.1, 78.7]} & 348     & 109 & 3.2:1  & 11.1 \tiny{[10.0, 12.3]} \\
  &                   & CoVe           & 75.0 & 73.9 \tiny{[73.1, 74.8]}          & 143     & 159 & 0.90:1 & 4.6 \tiny{[3.8, 5.3]}    \\
  &                   & Self-Refine    & 75.0 & 74.6 \tiny{[73.7, 75.4]}          & 6       & 35  & 0.17:1 & 0.2 \tiny{[0.1, 0.4]}    \\
\bottomrule
\end{tabular}
\caption{Same-model main results across our three primary benchmarks. Init/Final are accuracies (\%) for BBM and GPQA, and token-level F1 (\%) for HotpotQA. Imp./Deg.\ are example-level counts (an example is ``Improved'' if its correctness flipped from incorrect to correct, where correctness is exact match for BBM and GPQA and EM for HotpotQA). I:D = Imp / Deg, a precision-style ratio. Repair = Imp / N\textsubscript{wrong init} as a percentage, the recall-style ratio. Bold marks the highest value in each of Final, I:D, and Repair within each benchmark. Bracketed values are 95\% bootstrap confidence intervals (1{,}000 resamples) on final accuracy/F1 and on Repair rate.}
\label{tab:main-results}
\end{table*}
\endgroup

\paragraph{}DISC achieves the highest final accuracy or F1 on each of the three datasets in Table~\ref{tab:main-results}. At the per-cell level, DISC wins four of the five informative cells\footnote{Cell counts exclude GPQA gpt-4.1-nano, where the gate is inert across all methods (see Finding 3 in Section~\ref{sec:gpqa-fiveconfig}).}; BBM GPT-4o is the exception, where CoVe is higher in accuracy at substantially lower I:D. In three cells (BBM Sonnet 4.5, GPQA gpt-5.2, HotpotQA GPT-4o), DISC leads on both I:D and repair rate. In the remaining two (BBM GPT-4o, HotpotQA Sonnet 4.5), CoVe achieves higher repair at substantially lower I:D — CoVe attempts more corrections, fixes more in absolute terms, and breaks proportionally more correct answers.

\paragraph{BIG-Bench Mistake.} Baseline accuracy on the pre-generated traces is ${\sim}15\%$, leaving substantial headroom for correction. DISC's I:D on Sonnet 4.5 (149:1) is roughly $5\times$ that of the second-best baseline, Self-Refine (30:1), and roughly $13\times$ that of CoVe (11:1). Per-task results (Appendix~\ref{sec:appendix-bbm-pertask}) show DISC achieves zero degradation on four of five Sonnet tasks, while CoVe regresses substantially on multistep arithmetic and word sorting and Self-Refine yields net degradation on multistep arithmetic on both Sonnet and GPT-4o.

\paragraph{HotpotQA.} On HotpotQA's full validation set ($N{=}7{,}405$), DISC achieves the highest final F1 in both same-model configurations and is the only method that produces net F1 improvement on GPT-4o. The model-dependent pattern reflects a formatting-vs-reasoning split: Sonnet's verbose answer formatting lowers initial EM and F1, so simpler correction methods can recover formatting and catch up on absolute gain. EM, precision, and recall by method appear in Appendix~\ref{sec:appendix-hotpotqa}; a representative case is in Appendix~\ref{sec:appendix-examples}.

\paragraph{Compute.} The three methods differ in per-example cost. DISC issues one verification plan, 3--10 verification answers (depending on dataset difficulty), one judge call, and one corrector call when the judge fires \textsc{Mistake}, repeated up to $K{=}3$ iterations with early stopping. CoVe is single-pass: one plan, 3--10 verification answers, one revision. Self-Refine alternates feedback and refinement until convergence. Table~\ref{tab:compute} reports measured mean LLM API calls per example. DISC's call count is comparable to CoVe's on HotpotQA (gate frequently terminates early) and higher on BBM and GPQA Diamond (gate iterates more often). Self-Refine is consistently the cheapest (1--3 calls) because its critique step is monolithic; this cost advantage coincides with substantially lower performance on every benchmark in Table~\ref{tab:main-results}.

\begin{table}[t]
\centering
\small
\begin{tabular}{llccc}
\toprule
\textbf{Bench} & \textbf{Model} & \textbf{DISC} & \textbf{CoVe} & \textbf{Self-Ref.} \\
\midrule
HotpotQA & Sonnet 4.5 & 10.81   & 9.98    & 1.70 \\
HotpotQA & GPT-4o     & 9.45    & 9.90    & 1.03 \\
GPQA     & nano       & 11.05   & 6.88    & 1.33 \\
GPQA     & gpt-5.2    & 15.51   & 7.01    & 2.27 \\
BBM      & Sonnet 4.5 & 15.21   & 10.09   & 2.77 \\
BBM      & GPT-4o     & 15.87   & 9.87    & 2.35 \\
\bottomrule
\end{tabular}
\caption{Mean LLM API calls per example.}
\label{tab:compute}
\end{table}

\subsection{Cross-Model Configurations and Self-Confirmation Bias}
\label{sec:cross-model}

The same-model results in Section~\ref{sec:main-results} suggest that verification capability is a binding constraint: when the same model produces and verifies an answer, errors and verification evidence share systematic biases. We test this directly by varying the strength of the models that fill each pipeline role.

\paragraph{Five-configuration experiment on GPQA Diamond.}
\label{sec:gpqa-fiveconfig}
GPQA Diamond \citep{rein2023gpqa} consists of 198 graduate-level science questions. We vary which model fills each pipeline role between gpt-4.1-nano (weak) and gpt-5.2 (strong), yielding the five configurations summarized in Table~\ref{tab:gpqa-fiveconfig}.

\begin{table*}[t]
\centering
\footnotesize
\setlength{\tabcolsep}{3.5pt}
\begin{tabular}{lllllcccccc}
\toprule
\textbf{Config} & \textbf{Gen.} & \textbf{Verifier} & \textbf{Judge} & \textbf{Corr.} & \textbf{Init Acc} & \textbf{Final Acc} & \textbf{Imp} & \textbf{Deg} & \textbf{Net} & \textbf{Early Conv.} \\
\midrule
All nano & nano & nano & nano & nano & 54.0 \tiny{[47.5, 61.1]} & 54.0 \tiny{[47.5, 61.1]} & 0 & 0 & 0 & 100\% \\
Strong judge only & nano & nano & 5.2 & nano & 54.0 \tiny{[47.5, 61.1]} & 55.6 \tiny{[48.5, 62.6]} & 24 & 21 & $+3$ & 82.8\% \\
Strong verifier only & nano & 5.2 & nano & nano & 54.3 \tiny{[47.7, 61.4]} & 54.3 \tiny{[47.7, 61.4]} & 0 & 0 & 0 & 100\% \\
All gpt-5.2 & 5.2 & 5.2 & 5.2 & 5.2 & 75.3 \tiny{[69.2, 80.8]} & 79.3 \tiny{[73.2, 84.8]} & 18 & 10 & $+8$ & 84.3\% \\
Strong verify+judge & nano & 5.2 & 5.2 & nano & 54.3 \tiny{[47.2, 61.4]} & 71.6 \tiny{[65.0, 77.7]} & 51 & 17 & $+34$ & 78.7\% \\
\bottomrule
\end{tabular}
\caption{GPQA Diamond: five-configuration cross-model experiment. Init Acc and Final Acc are accuracies (\%) with 95\% bootstrap CIs in brackets (1{,}000 resamples). All configurations use the same pre-generated initial responses (per generator); 5.2 = gpt-5.2, nano = gpt-4.1-nano. Net = Imp $-$ Deg. Example counts vary slightly across rows (197--198) due to API content-policy refusals on individual questions.}
\label{tab:gpqa-fiveconfig}
\end{table*}

Three findings stand out from this comparison. 1.Self-confirmation bias is the primary obstacle to same-model self-correction. With identical verifier and judge models, the only difference between the all-gpt-5.2 row and the cross-model row is whose outputs are being checked. The gpt-5.2 verifier surfaces 51 errors in nano's outputs against 18 in its own, consistent with the systematic-bias account of intrinsic self-correction \citep{huang2024large,kamoi2024when}: a model's verification reflects the same priors that produced its errors. The judge's classifier statistics tell the same story (Appendix~\ref{sec:appendix-confusion}): the gpt-5.2 judge has 71.3\% precision and 85.6\% recall on nano's outputs but drops to 62.0\% precision and 63.3\% recall on its own. This aligns with \citet{zhang2024small}'s finding that effective self-correction depends on strong external verification. DISC is ensemble-like in design, aggregating evidence through a judge; cross-model role allocation improves both ingredients by introducing a stronger verifier and reducing shared systematic errors between generation and verification.

2. Verifier-judge synergy is strongly super-additive. Strong judge alone, $+3$ net; strong verifier alone, $0$ net; both, $+34$. The interaction effect is roughly an order of magnitude larger than the sum of the parts.

3. Weak judges do not activate correction.
\label{sec:rationale-analysis}
The nano judge said \textsc{No\_Mistake} on 100\% of all 197 examples in the strong-verifier-only configuration, including all 90 wrong answers; the pipeline never activated a correction. The same gpt-5.2 verification evidence enables the gpt-5.2 judge in the cross-model configuration to flag 86\% of errors (Appendix~\ref{sec:appendix-confusion}), localizing the bottleneck to judge reasoning rather than evidence quality. This is consistent with \citet{zhang2024small}'s finding that small models need strong external verification. This experiment shows that even strong evidence cannot compensate for a weak judge.

To investigate this, we classify the 90 wrong-answer cases along two dimensions, using Claude Opus 4.5 as an LLM-as-judge classifier (full procedure in Appendix~\ref{sec:appendix-rationale-method}): \emph{verification classification} labels whether the verifier's evidence contradicts the chosen answer, and \emph{rationale classification} labels whether the nano judge's rationale acknowledges that contradiction before concluding \textsc{No\_Mistake}. The verifier produces contradicting evidence in 89 of 90 cases, yet the nano judge concludes \textsc{No\_Mistake} every time. The classifier splits the 89 cases into two failure modes: in 48 of 89 the judge does not acknowledge the contradiction, treating verification as confirmatory regardless of content (\emph{Blindness}); in 41 of 89 the judge surfaces the contradiction in language but still concludes \textsc{No\_Mistake} (\emph{Inability to act}). Appendix~\ref{sec:appendix-examples} gives qualitative examples.

\subsection{Ablations on DISC}
\label{sec:ablations}

\paragraph{Isolating the Judgment Gate.}
\label{sec:gate-ablation}
The architectural differences between DISC and CoVe are the judgment gate and prompt design. We address them with two complementary ablations: a forced-\textsc{Mistake} ablation that varies the gate while holding DISC's prompts fixed (this section), and a prompt-swap ablation that varies the prompts while holding the gateless single-pass architecture fixed. In the gate ablation, holding the verifier, corrector, and all prompts fixed, the judge produces a brief \textsc{Mistake}-style rationale on every example regardless of verification evidence, and the corrector runs unconditionally.

\begin{table*}[!ht]
\centering
\small
\begin{tabular}{lllcccrc}
\toprule
\textbf{Benchmark} & \textbf{Model} & \textbf{Method} & \textbf{Final} & \textbf{Imp} & \textbf{Deg} & \textbf{Net} & \textbf{I:D} \\
\midrule
BBM Dyck Languages & Sonnet 4.5 & DISC         & 50.8 & 221  & 8   & $+213$  & 27.6:1 \\
                   &            & DISC-NoGate  & 47.1 & 222  & 31  & $+191$  & 7.2:1  \\
BBM Dyck Languages & GPT-4o     & DISC         & 18.9 & 41   & 17  & $+24$   & 2.4:1  \\
                   &            & DISC-NoGate  & 13.2 & 44   & 54  & $-10$   & 0.81:1 \\
\midrule
HotpotQA           & Sonnet 4.5 & DISC         & 64.1 & 1487 & 109 & $+1378$ & 13.6:1 \\
                   &            & DISC-NoGate  & 62.3 & 1812 & 568 & $+1244$ & 3.2:1  \\
HotpotQA           & GPT-4o     & DISC         & 60.9 & 348  & 109 & $+239$  & 3.2:1  \\
                   &            & DISC-NoGate  & 62.1 & 605  & 278 & $+327$  & 2.2:1  \\
\bottomrule
\end{tabular}
\caption{Forced-\textsc{Mistake} ablation: removing the judgment gate while holding the verifier, corrector, and all prompts fixed. All runs at $K{=}1$. Final is accuracy on BBM Dyck Languages ($N{=}592$) and EM on HotpotQA ($N{\approx}7{,}400$). Initial accuracy: 14.9\% on BBM Dyck Languages (both models share the same pre-generated trace); initial EM 45.5\% (Sonnet 4.5) and 57.7\% (GPT-4o) on HotpotQA.}
\label{tab:gate-ablation}
\end{table*}

Two findings stand out. First, I:D drops 2--4$\times$ in every cell, with degradation roughly doubling on Sonnet and tripling on GPT-4o, isolating the gate's contribution as blocking rewrites that damage already-correct answers. Second, the verifier-corrector pass still produces substantial recovery without the gate, showing that recall and precision are architecturally separable. The gate's effect on net accuracy is configuration-dependent: on Dyck Languages it is uniformly net-positive; on HotpotQA the gate gains $+134$ EM on Sonnet but loses $88$ EM on GPT-4o, because the conservative gate blocks more would-be-improvements than would-be-degradations (see Limitations).

\paragraph{Prompt Ablation.}
\label{sec:prompt-ablation}
The forced-\textsc{Mistake} ablation isolates the gate while holding DISC's prompts fixed, but it does not address the prompt-language confound: DISC's verification and correction prompts differ from CoVe's, and the gap in Table~\ref{tab:main-results} could in principle reflect any of them. To test whether DISC's prompts alone account for its gains, we plug DISC's verification, evidence, and correction prompts into CoVe's single-pass gateless architecture, holding all other variables fixed. On GPQA Diamond with gpt-5.2 ($N{=}198$), this prompt-swap ablation reaches 67.2\% final accuracy, below both native CoVe (69.7\%) and DISC (79.3\%); full results are in Appendix~\ref{sec:appendix-prompt-ablation}. DISC's prompts continue to fix errors at a rate comparable to DISC, but without the gate the corrector also rewrites already-correct answers as if they contained a confirmed error. 

\paragraph{Iteration Budget.}
\label{sec:ablation-iter}
We sweep $K \in \{1, 2, 3, 5, 7, 8\}$ on GPQA Diamond in the cross-model configuration (gpt-4.1-nano generator and corrector, gpt-5.2 verifier and judge). $K{=}1{\to}K{=}3$ produces the bulk of the gain: net rises from $+22$ to $+34$ and I:D from 2.0:1 to 3.0:1. From $K{=}3$ onward the curve plateaus (slight dip to $+33$ at $K{=}5$, recovery to $+35$ at $K{=}8$), and $K{=}3$ captures 97\% of the maximum net gain at 70\% of the average iteration count (1.86 vs.\ 2.63). Cross-model DISC at $K{=}1$ ($+22$) already outperforms same-model gpt-5.2 DISC at $K{=}3$ ($+8$; Table~\ref{tab:main-results}): judge quality matters more than iteration count. See Appendix~\ref{sec:appendix-iteration} for the full table and further discussion.

\section{Discussion}

A consistent pattern across our experiments is that the gains from DISC trace to architectural choices in the verification pipeline rather than to the raw capability of any single component. The judgment gate, the separation of the verifier from the generator, and the option to route different roles to different models each contribute to the observed correction quality, and the most reliable configurations are the ones in which these choices compound.

\subsection{The Judge as Noise Filter}

Without the gate, DISC would resemble CoVe with iteration: each round revises unconditionally and degradation can compound. The gate makes iteration \emph{safer}. Most rounds either improve the answer or leave it unchanged, consistent with CoVe's higher degradation rates in our same-model comparisons (Table~\ref{tab:main-results}). This is the mechanism behind DISC's consistent precision advantage: iterating doesn't help if each iteration risks new degradations as the price of attempted repair.

\subsection{When DISC Helps Most}
\label{sec:capability-floor}

Three regimes emerge from our experiments. First, when the same-model verifier and judge are capable enough to engage the gate, DISC produces clean preservation-priority gains. Second, when the same-model verifier or judge cannot engage the gate, DISC neither helps nor harms. Third, upgrading verification and judgment to a stronger model engages the gate
where the same-model version did not. 

A two-stage capability requirement underlies these regimes: the verifier must produce evidence accurate enough to act on, and the judge must be capable enough to translate that evidence into a \textsc{Mistake} flag. Failures at the judge stage are visible as the \emph{Blindness} and \emph{Inability to act} modes documented in Finding~3 (\S\ref{sec:gpqa-fiveconfig}): the evidence comes from a frontier model, but the nano judge can neither reliably read it nor act on it. This complements the feedback-bottleneck observations of \citet{huang2024large} and \citet{kamoi2024when}. Even with strong external feedback, the judgment step can fail. Failures at the verifier stage are harder to isolate in our experiments, since most cases where the verifier is wrong are also cases where the judge fails to flag the error. 

A milder form of the same-model underperformance appears across Table~\ref{tab:main-results}: same-model DISC produces large gains on benchmarks with structural slack (BBM Sonnet, where the input is a pre-generated trace; HotpotQA Sonnet, where verbose formatting lowers initial scores). GPQA Diamond removes both advantages and the questions sit at the model's capability frontier. Same-model gpt-5.2 DISC nets only $+8$ despite using the strongest model in our evaluation. A single gpt-5.2 call, however, already reaches 75.3\% on GPQA, above cross-model DISC's final accuracy. Therefore the strongest practical wins come from same-model DISC on benchmarks where one frontier-model call does not already suffice, such as BBM and HotpotQA. Our cross-model experiments paired weak generators with strong verifiers; strong-generator plus different-strong-verifier configurations, where the verifier's training distribution differs from the generator's, are untested and a natural next step.

\section{Conclusion}

DISC's contribution is architectural: a binary judgment gate,
iterative refinement over multiple passes, and modular roles that
can be filled by different models. The gate enables safe iteration; routing verification and judgment to a stronger model than the generator further extends correction to weaker generators. DISC outperforms single-pass and iterative baselines on the precision-recall trade-off when the verifier and judge can act on their evidence. A capability floor bounds this behavior: below it, the
judge cannot engage the gate and DISC is inert. Our rationale analysis decomposes this floor into two judge failure modes, \emph{Blindness} and \emph{Inability to act}. Two capability requirements of test-time self-correction follow from our analysis. The verifier must be accurate enough to surface evidence that contradicts an incorrect response; the judge must
then be capable enough both to read and act on that evidence.

\section*{Limitations}

\textbf{Prompt sensitivity and operating-point control.} The judgment gate is sensitive to prompt framing, as is typical of prompt-based components. The gate ablation (Section~\ref{sec:gate-ablation}) shows that the gate itself functions as a tunable operating-point control. Removing it uniformly shifts the operating point toward lower precision and higher recall. Two principled extensions of this knob are natural and we leave them to future work: explicit prompt sweeps that target a desired precision or recall level, and thresholding on the judge's $P(\textsc{Mistake})$ token probability, which would expose a continuous precision-recall control without modifying the prompt. 

\textbf{Verifiable ground truth.} All three benchmarks have objectively verifiable answers. DISC's applicability to open-ended tasks (summarization and creative writing) is unaddressed. The judgment gate assumes that ``correct'' and ``incorrect'' are well-defined; this assumption weakens for tasks with subjective or multi-faceted evaluation criteria.

\textbf{Operating-regime sensitivity.} DISC's judgment gate prioritizes precision: at low baselines where most answers are already wrong, an always-revise strategy can match or exceed gated correction on raw accuracy by attempting more rewrites, even when its correction precision (I:D) is substantially lower. The gate ablation (Section~\ref{sec:gate-ablation}) characterizes this trade-off directly. The gate always raises I:D, but its net effect on accuracy depends on judge calibration in the target regime: where the judge reliably separates good corrections from harmful ones (BBM Dyck on both models; HotpotQA on Sonnet~4.5) the gate is net-positive; where the judge is less reliable on the benchmark's error subset (HotpotQA on GPT-4o) the gate's operating point becomes too conservative and net accuracy is slightly higher without it ($+327$ vs.\ $+239$ EM). A complementary version of the same trade-off appears within BBM at a per-task level (DISC dominates Sonnet 4.5 across all tasks but trails CoVe on most GPT-4o tasks; Appendix~\ref{sec:appendix-bbm-pertask}). DISC's design assumes that preserving correct answers is at least as important as repairing incorrect ones; for applications where this assumption does not hold, a less conservative variant (e.g., a softer judgment threshold, the no-gate variant, or thresholding on judge $P(\textsc{Mistake})$) may be preferable.

\textbf{Future work.} Natural extensions include training-based judge calibration, adapting the framework to open-ended tasks, and grounding verification in external data such as search results for claims beyond the model's parametric knowledge.

\bibliography{custom}

\appendix

\section{Prompt Templates}
\label{sec:appendix-prompts}

As a worked example we reproduce below the full DISC prompt templates used in our BIG-Bench Mistake configuration (Section~\ref{sec:main-results}; BBM is the benchmark where DISC most decisively dominates the baselines, see Table~\ref{tab:main-results}). The same structural template -- verifier, judge, corrector, with the same input/output schemas -- is used for every benchmark; per-stage prompts are tailored to each benchmark's knowledge source, answer format, and judgment semantics. Curly-braced tokens (\texttt{\{question\}}, \texttt{\{response\}}, \texttt{\{verification\_question\}}, \texttt{\{verification\_results\}}) are placeholders filled in at runtime.

\begin{figure*}[!ht]
\begin{Verbatim}[fontsize=\small]
You are an expert fact-checker and logical reasoner. Your task is
to generate verification questions to validate a given response.

Given:
- Question: {question}
- Initial Response: {response}

Generate 3-5 targeted verification questions that will help check:
1. Factual accuracy of claims made
2. Logical consistency of the reasoning
3. Correctness of any calculations or processes
4. Validity of assumptions

Format your output as a numbered list of verification questions.
Each question should be:
- Specific and focused on one aspect
- Answerable independently
- Designed to reveal potential errors

Verification Questions:
\end{Verbatim}
\caption{Verification question generation prompt. The requested question-count range is benchmark-specific: 3--5 for BBM and HotpotQA, 4--10 for GPQA Diamond, reflecting differences in task difficulty.}
\label{fig:prompt-verification}
\end{figure*}

\begin{figure*}[!ht]
\begin{Verbatim}[fontsize=\small]
You are an expert fact-checker to verify a response to an
academic reasoning task. Answer the verification question
carefully and accurately.

Original Question: {question}

Verification Question: {verification_question}

IMPORTANT:
- If the question involves counting, count twice to verify your
  count is correct
- Show your work step-by-step for any calculations or counting
- If uncertain about any claim, acknowledge uncertainty rather
  than making false assertions
- Double-check any conclusion about whether the answer is wrong

Provide a clear, accurate answer to this verification question.

Answer:
\end{Verbatim}
\caption{Evidence generation prompt. The HotpotQA variant additionally requires the verifier to quote the exact answer phrase from the provided context.}
\label{fig:prompt-evidence}
\end{figure*}

\begin{figure*}[!ht]
\begin{Verbatim}[fontsize=\small]
You are an expert at identifying logical mistakes in reasoning
traces.

Original Question: {question}

Original Response with steps:
{response}

Verification Q&A Results:
{verification_results}

Task: Based on the verification Q&A, identify ALL steps (0-indexed)
that contain mistakes.

Important notes:
- The response has steps formatted as "Thought 0:", "Thought 1:",
  "Thought 2:", etc.
- Each thought number corresponds directly to the step index
  (Thought 0 = step 0, Thought 1 = step 1, etc.)
- Only identify steps with LOGICAL MISTAKES (wrong reasoning,
  incorrect calculations, false assumptions)
- Do NOT flag steps just for being unclear or poorly formatted
- If a mistake in an early step causes errors in later steps,
  list ALL affected steps

Your response MUST be in ONE of these exact formats:
1. If NO mistakes found: "No mistakes"
2. If mistakes found:
   "Mistake steps: [list of 0-indexed step numbers]"

Examples:
- "No mistakes"
- "Mistake steps: 0"          (mistake only in Thought 0)
- "Mistake steps: 0, 3, 4"    (mistakes in Thoughts 0, 3, and 4)
\end{Verbatim}
\caption{Judge prompt. On BIG-Bench Mistake the judge returns a list of mistaken step indices and the gate fires when the list is non-empty; on GPQA Diamond and HotpotQA the judge returns a binary \textsc{Mistake}/\textsc{No\_Mistake} decision with a free-text rationale.}
\label{fig:prompt-judgment}
\end{figure*}

\begin{figure*}[!ht]
\begin{Verbatim}[fontsize=\small]
You are an expert problem solver. Based on verification results,
correct any errors in the initial response.

Original Question: {question}
Initial Response: {initial_response}

Verification Results:
{verification_results}

Based on the verification results above, provide a corrected
response that fixes any identified errors.

Important:
- Maintain the EXACT same format as the initial response
- If the initial response has "Thought 0:", "Thought 1:", etc.,
  your corrected response MUST also have this format
- Fix any incorrect reasoning, calculations, or conclusions
- Keep correct parts unchanged
- End with "The answer is [your answer]" on a new line

Format your corrected response EXACTLY like the initial response
format.

Corrected Response:
\end{Verbatim}
\caption{Correction prompt. The HotpotQA variant outputs a short answer span; the GPQA Diamond variant outputs a multiple-choice letter.}
\label{fig:prompt-correction}
\end{figure*}

\section{Model Settings}
\label{sec:appendix-model-settings}

\paragraph{Model selection.} The four models span a range of capabilities: two current-generation frontier LLMs (Claude Sonnet 4.5, gpt-5.2), an established previous-generation model (GPT-4o), and a small model (gpt-4.1-nano). This range lets us test DISC under both same-model and cross-model configurations. The BIG-Bench Mistake and HotpotQA evaluations are same-model on Sonnet 4.5 and on GPT-4o. The GPQA Diamond pairing of gpt-4.1-nano (generator) and gpt-5.2 (verifier and judge) creates the capability gap needed to isolate self-confirmation bias from generator capability.

\paragraph{Reasoning and thinking settings.} We disable built-in reasoning/thinking modes for all models. For Claude Sonnet 4.5, extended thinking is not enabled. For gpt-5.2, \texttt{reasoning\_effort} is left at its default value, which does not invoke explicit reasoning. GPT-4o and gpt-4.1-nano do not expose a comparable reasoning mode.

\section{BIG-Bench Mistake Per-Task Results}
\label{sec:appendix-bbm-pertask}

\subsection*{Claude Sonnet 4.5}

\begin{table*}[!ht]
\centering
\footnotesize
\setlength{\tabcolsep}{3pt}
\begin{tabular}{lrccc}
\toprule
\textbf{Task} & \textbf{N} & \textbf{DISC} & \textbf{CoVe} & \textbf{Self-Refine} \\
\midrule
Tracking Shuffled Objects & 300 & \textbf{100.0}                       & 99.0 \tiny{[97.7, 100.0]} & \textbf{100.0} \\
Logical Deduction         & 300 & \textbf{99.3} \tiny{[98.3, 100.0]}   & 97.7 \tiny{[95.7, 99.3]}  & 98.3 \tiny{[96.7, 99.7]} \\
Multistep Arithmetic      & 300 & \textbf{98.3} \tiny{[96.7, 99.7]}    & 43.7 \tiny{[38.0, 49.0]}  & 15.3 \tiny{[11.7, 20.0]} \\
Word Sorting              & 300 & \textbf{89.6} \tiny{[86.2, 93.0]}    & 61.1 \tiny{[55.7, 66.2]}  & 53.2 \tiny{[47.2, 58.9]} \\
Dyck Languages            & 592 & \textbf{50.8} \tiny{[47.0, 54.9]}    & 24.5 \tiny{[21.3, 27.7]}  & 15.0 \tiny{[12.5, 18.1]} \\
\bottomrule
\end{tabular}
\caption{BIG-Bench Mistake per-task accuracy (Claude Sonnet 4.5). Bracketed values are 95\% bootstrap CIs (1{,}000 resamples); CIs omitted for cells at 100\% (zero variance). Bold marks the best accuracy in each row. Per-task degradation rates: DISC produces 0\% degradation on four of five tasks (9.1\% on Dyck Languages); CoVe degrades 2.2\% on Tracking Shuffled Objects, 28.9\% on Multistep Arithmetic, 35.6\% on Word Sorting, and 51.1\% on Dyck Languages; Self-Refine degrades 17.0\% on Dyck Languages, 4.4\% on Multistep Arithmetic, 9.1\% on Word Sorting, and 0.0\% on the remaining tasks. The $N$ column reports the canonical dataset size; per-method effective $N$ on Word Sorting was 298 (DISC), 296 (CoVe), and 282 (Self-Refine) due to API content-policy blocks on individual examples.}
\label{tab:bbm-pertask-sonnet}
\end{table*}

DISC achieves near-perfect accuracy on three of five tasks (tracking shuffled objects, logical deduction, multistep arithmetic) with zero degradation on four of five. CoVe shows substantial regressions on multistep arithmetic (43.7\% accuracy, 28.9\% degradation) and word sorting (61.1\%, 35.6\%): without a judgment gate, revision applies unconditionally. Self-Refine performs well on the two tasks where the verification step is essentially trivial (Tracking Shuffled Objects, Logical Deduction) but collapses on Multistep Arithmetic (15.3\%, only 1.2\% improvement) and Dyck Languages (15.0\%, 3.5\% improvement): self-feedback cannot reliably detect calculation errors or formal pattern violations.

\subsection*{GPT-4o}

\begin{table*}[!ht]
\centering
\footnotesize
\setlength{\tabcolsep}{3pt}
\begin{tabular}{lrccc}
\toprule
\textbf{Task} & \textbf{N} & \textbf{DISC} & \textbf{CoVe} & \textbf{Self-Refine} \\
\midrule
Tracking Shuffled Objects & 300 & 92.0 \tiny{[89.0, 94.7]} & \textbf{97.3} \tiny{[95.3, 99.0]} & 68.3 \tiny{[63.0, 73.3]} \\
Logical Deduction         & 300 & 65.0 \tiny{[59.7, 70.7]} & \textbf{81.0} \tiny{[76.7, 85.3]} & 55.7 \tiny{[50.0, 61.0]} \\
Multistep Arithmetic      & 300 & 54.0 \tiny{[48.3, 59.3]} & \textbf{95.0} \tiny{[92.7, 97.0]} & 14.0 \tiny{[10.3, 18.0]} \\
Word Sorting              & 300 & 50.0 \tiny{[44.3, 55.3]} & \textbf{75.9} \tiny{[70.9, 80.6]} & 13.3 \tiny{[9.3, 17.0]}  \\
Dyck Languages            & 592 & \textbf{18.9} \tiny{[16.0, 22.1]} & 10.8 \tiny{[8.3, 13.3]} & 6.8 \tiny{[4.9, 9.0]}    \\
\bottomrule
\end{tabular}
\caption{BIG-Bench Mistake per-task accuracy (GPT-4o). Bracketed values are 95\% bootstrap CIs (1{,}000 resamples). Per-task degradation rates: DISC produces 0\% degradation on four of five tasks (19.3\% on Dyck Languages); CoVe degrades 6.7\% on Multistep Arithmetic, 4.4\% on Word Sorting, 2.2\% on Tracking Shuffled Objects, 2.2\% on Logical Deduction, and 44.3\% on Dyck Languages; Self-Refine degrades 54.5\% on Dyck Languages, 6.7\% on Multistep Arithmetic, 11.1\% on Word Sorting, and 0\% on both Tracking Shuffled Objects and Logical Deduction. The $N$ column reports the canonical dataset size; per-method effective $N$ on Word Sorting was 300 (DISC), 299 (CoVe), and 300 (Self-Refine) due to API content-policy blocks on individual examples.}
\label{tab:bbm-pertask-gpt4o}
\end{table*}

With GPT-4o, CoVe outperforms DISC on most tasks (except Dyck languages) in raw accuracy. However, DISC achieves lower degradation on every task. GPT-4o's weaker verification capability means DISC's conservative judgment gate fires less often, limiting the number of corrections attempted. This reversal highlights the interaction between model capability and method design: DISC requires sufficient model capability to \emph{detect} mistakes through verification; without it, the judgment gate correctly but unhelpfully preserves the (often incorrect) status quo. Self-Refine collapses on every task and is catastrophic on Dyck Languages, where it degrades 54.5\% of originally correct answers and improves zero, indicating that GPT-4o's structured self-feedback is actively harmful for formal pattern tasks.

\section{Iteration Budget Sweep: Full Results}
\label{sec:appendix-iteration}

Section~\ref{sec:ablation-iter} summarizes the iteration sweep on GPQA Diamond. The full per-K counts and convergence statistics are reproduced below.

\begin{table*}[!ht]
\centering
\small
\begin{tabular}{cccccccccc}
\toprule
\textbf{K} & \textbf{N} & \textbf{Init Acc} & \textbf{Final Acc} & \textbf{Improved} & \textbf{Degraded} & \textbf{Net} & \textbf{I:D} & \textbf{Avg Iter} & \textbf{Converged} \\
\midrule
1 & 198 & 54.0\% & 65.2\% & 45 & 23 & $+22$ & 2.0:1 & 1.00 & 46.5\% \\
2 & 198 & 54.0\% & 67.7\% & 51 & 24 & $+27$ & 2.1:1 & 1.51 & 70.7\% \\
3 & 197 & 54.0\% & 71.6\% & 51 & 17 & $+34$ & 3.0:1 & 1.86 & 78.7\% \\
5 & 197 & 54.0\% & 70.6\% & 51 & 18 & $+33$ & 2.8:1 & 2.36 & 83.8\% \\
7 & 195 & 54.0\% & 72.3\% & 50 & 16 & $+34$ & 3.1:1 & 2.69 & 89.2\% \\
8 & 197 & 54.0\% & 72.1\% & 51 & 16 & $+35$ & 3.2:1 & 2.63 & 90.4\% \\
\bottomrule
\end{tabular}
\caption{GPQA Diamond cross-model iteration budget sweep (gpt-4.1-nano generator and corrector, gpt-5.2 verifier and judge). All runs share the same pre-generated initial responses (54.0\% baseline). $N$ varies slightly across $K$ values due to API content-policy refusals on individual questions. ``Avg Iter'' is the mean number of iterations per example before convergence; ``Converged'' is the fraction of examples on which the judge returned \textsc{No\_Mistake} at or before iteration $K$. The $K{=}3$ row corresponds to the cross-model configuration in Table~\ref{tab:gpqa-fiveconfig}.}
\label{tab:iteration-sweep}
\end{table*}

\paragraph{Three phases.} $K{=}1{\to}K{=}3$ is the rapid-improvement regime: net rises $+22{\to}+34$ as degradation falls $23{\to}17$ and I:D climbs from 2.0:1 to 3.0:1. $K{=}3{\to}K{=}5$ slightly regresses ($+33$, 2.8:1) as additional iterations introduce one new degradation without new improvements. $K{=}5{\to}K{=}8$ gradually recovers to a marginal further improvement at $K{=}8$ ($+35$, 3.2:1). $K{=}3$ remains the sweet spot: it captures 97\% of the maximum net gain at 70\% of the average iteration cost.

\paragraph{Degradation does not compound with K.} The degradation count drops as $K$ rises: 23 ($K{=}1$) $\to$ 17 ($K{=}3$) $\to$ 16 ($K{=}7$ and $K{=}8$). More iterations do not increase the risk of breaking correct answers, consistent with the gate's preservation role: some answers degraded in early iterations are recovered by later iterations when the verifier produces additional or corrected evidence.

\paragraph{Convergence.} At $K{=}1$, only 46.5\% of examples converge (judge says \textsc{No\_Mistake} on the first pass). Convergence rises to 78.7\% at $K{=}3$ and to 90.4\% at $K{=}8$. The remaining $\sim$10\% of examples represent persistent disagreements where the judge keeps flagging \textsc{Mistake} through all available iterations; the long tail is real (21 examples exhaust all 8 iterations at $K{=}8$) but its contribution to net correction at $K{=}8$ vs $K{=}3$ is small.

\paragraph{Comparison to same-model.} Cross-model DISC at $K{=}1$ achieves $+22$ net, already higher than same-model gpt-5.2 DISC at $K{=}3$ ($+8$; Table~\ref{tab:main-results}). Iteration cannot substitute for verifier and judge capability: at the canonical $K{=}3$ budget, the same-model setting produces less than a quarter of the net correction that the cross-model setting produces at one iteration.

\section{HotpotQA: Per-Metric Results}
\label{sec:appendix-hotpotqa}

Table~\ref{tab:hotpotqa-detail} reports the full HotpotQA results across the four answer-quality metrics (EM, F1, precision, recall). Table~\ref{tab:main-results} in the body reports F1 as the primary metric, since HotpotQA's reference answers are short free-text spans for which token-level F1 is the standard scoring metric (robust to formatting variation and partial overlap); the additional metrics here -- EM, precision, and recall -- expose the components of F1 and the formatting gap discussed in the comparability note in Section~\ref{sec:main-results}. Precision and recall in this appendix follow the standard SQuAD / HotpotQA evaluation: \emph{token-level precision} is the fraction of predicted answer tokens that appear in the ground-truth answer, and \emph{token-level recall} is the fraction of ground-truth tokens that appear in the prediction; F1 is their harmonic mean. These are within-example answer-quality metrics and are distinct from the I:D ratio and repair rate in Table~\ref{tab:main-results}, which are across-example correction-outcome metrics (precision and recall of the correction process itself). All DISC entries use $K{=}3$.

\begin{table*}[!ht]
\centering
\footnotesize
\setlength{\tabcolsep}{3pt}
\begin{tabular}{llcccc}
\toprule
\textbf{Model} & \textbf{Method} & \textbf{EM} & \textbf{F1} & \textbf{Prec.} & \textbf{Rec.} \\
\midrule
Sonnet 4.5 & Initial     & 45.5 \tiny{[44.3, 46.5]}          & 65.1 \tiny{[64.2, 65.9]}          & 63.0 \tiny{[62.0, 63.9]}          & 83.9 \tiny{[83.2, 84.6]} \\
Sonnet 4.5 & DISC        & 64.1 \tiny{[62.9, 65.1]}          & \textbf{80.4} \tiny{[79.7, 81.1]} & 81.6 \tiny{[80.8, 82.3]}          & 85.7 \tiny{[85.0, 86.4]} \\
Sonnet 4.5 & CoVe        & \textbf{65.2} \tiny{[64.1, 66.3]} & 80.0 \tiny{[79.3, 80.8]}          & \textbf{82.4} \tiny{[81.6, 83.1]} & 83.4 \tiny{[82.7, 84.1]} \\
Sonnet 4.5 & Self-Refine & 59.7 \tiny{[58.5, 60.9]}          & 77.8 \tiny{[77.0, 78.5]}          & 78.0 \tiny{[77.1, 78.7]}          & \textbf{86.5} \tiny{[85.8, 87.1]} \\
\midrule
GPT-4o     & Initial     & 57.7 \tiny{[56.5, 58.8]}          & 75.0 \tiny{[74.2, 75.8]}          & 75.9 \tiny{[75.0, 76.7]}          & 79.7 \tiny{[78.8, 80.5]} \\
GPT-4o     & DISC        & \textbf{60.9} \tiny{[59.8, 62.0]} & \textbf{77.9} \tiny{[77.1, 78.7]} & \textbf{78.8} \tiny{[78.0, 79.7]} & \textbf{82.3} \tiny{[81.5, 83.1]} \\
GPT-4o     & CoVe        & 57.5 \tiny{[56.3, 58.6]}          & 73.9 \tiny{[73.1, 74.8]}          & 75.0 \tiny{[74.1, 75.8]}          & 78.7 \tiny{[77.9, 79.5]} \\
GPT-4o     & Self-Refine & 57.3 \tiny{[56.2, 58.4]}          & 74.6 \tiny{[73.7, 75.4]}          & 75.4 \tiny{[74.5, 76.3]}          & 79.3 \tiny{[78.5, 80.2]} \\
\bottomrule
\end{tabular}
\caption{HotpotQA per-metric results across same-model configurations and methods on the full validation set ($N{=}7{,}405$). ``Initial'' rows report the uncorrected baseline; remaining rows report post-method scores. Bold entries mark the best score in each (model, metric) cell. Bracketed values are 95\% bootstrap CIs (1{,}000 resamples).}
\label{tab:hotpotqa-detail}
\end{table*}

Two patterns are worth flagging. On Sonnet 4.5, all three correction methods produce large precision gains ($+15$ to $+19$pp) but only small recall changes ($+1.8$, $-0.5$, $+2.6$pp): the initial answers already contain the right tokens but bury them in extra words, so correction primarily trims rather than recovers content. CoVe achieves the highest final precision (82.4\%) at the cost of slightly hurt recall ($-0.5$pp), the only Sonnet method whose recall does not improve. On GPT-4o, where there is less formatting slack to begin with, only DISC improves all three metrics simultaneously ($+2.96$pp precision, $+2.60$pp recall, $+2.90$pp F1); CoVe and Self-Refine each degrade all three. EM in Table~\ref{tab:main-results} is the strictest of these metrics; the F1, precision, and recall numbers here show the same model-dependent pattern (Sonnet correction is largely formatting cleanup, GPT-4o correction requires reasoning) without EM's binary cliff.

\section{Qualitative Examples}
\label{sec:appendix-examples}

We present two qualitative examples that illustrate different DISC behaviors: a HotpotQA case showing formatting recovery on Sonnet 4.5, and a GPQA Diamond case in the strong-verifier failure regime where the weak nano judge acknowledges the contradiction surfaced by verification but does not act on it (cf.\ Section~\ref{sec:gpqa-fiveconfig}, Finding 3).

\paragraph{Example (HotpotQA formatting recovery: Sonnet 4.5 same-model).}
Question: \emph{``Which New South Wales town is larger, Port Macquarie or Bonny Hills?''} (reference answer: \emph{``Port Macquarie''}). Sonnet's initial response carries the correct reasoning but embeds the answer in a full sentence: \emph{``Port Macquarie is larger than Bonny Hills, with a population of 45{,}698 compared to Bonny Hills' population of 2{,}870.''} The extra tokens score EM=0 and token-level F1=0.18 (recall 1.0, precision 0.11) despite the underlying answer being correct. After DISC, the final answer corrects to \emph{``Port Macquarie''} (EM=1, F1=1.0). GPT-4o's initial response on the same example is already \emph{``Port Macquarie''}.

\paragraph{Example (strong-verifier failure: nano gen + gpt-5.2 verify, nano judge).}
The gpt-5.2 verifier states that O$_2$ is \emph{``thermodynamically a weaker oxidant in basic solutions,''} directly contradicting option~D. The nano judge writes: \emph{``which contradicts the thermodynamic aspects supporting option D''}, then continues: \emph{``However\ldots\ the key points are consistent with option D\ldots\ Therefore, the answer does not need correction.''} The judge surfaces the contradiction in its own rationale before concluding \textsc{No\_Mistake}; this illustrates the \emph{inability to act} failure mode (\S\ref{sec:capability-floor}), which an LLM-as-judge classification matches in 41 of 89 (46.1\%) wrong-answer cases where the verifier produces contradicting evidence. The remaining 48 of 89 (53.9\%) are \emph{Blindness} cases in which the rationale does not engage with the verifier's contradicting evidence at all and instead frames it as confirmatory.

\section{Judge Rationale Analysis: Methodology}
\label{sec:appendix-rationale-method}

This appendix gives the full procedure used in Section~\ref{sec:rationale-analysis} and Section~\ref{sec:gpqa-fiveconfig} (Finding 3).

\paragraph{Inputs.} Run~3 of the GPQA Diamond cross-model experiment (nano generator + corrector, gpt-5.2 verifier, nano judge; see Table~\ref{tab:gpqa-fiveconfig}) yields 90 wrong-answer cases on which the nano judge concluded \textsc{No\_Mistake}. For each case the analysis takes (i) the question and answer choices, (ii) the chosen wrong answer, (iii) the ground-truth correct answer, (iv) the verifier's question-answer pairs, and (v) the judge's natural-language rationale.

\paragraph{Two classification modes.} Each case is classified along two independent dimensions.

\emph{Verification mode} asks whether the verifier surfaced contradicting evidence at all. The classifier sees the question, the chosen wrong answer, the correct answer, and the verifier Q\&A pairs, and labels the evidence as \textsc{Contradicts} or \textsc{Does\_Not\_Contradict}. A label of \textsc{Contradicts} requires that the evidence contains facts, calculations, or reasoning inconsistent with the chosen answer or supportive of the correct answer; the contradiction may be implicit (e.g., the evidence derives a value matching a different option) or explicit. \textsc{Does\_Not\_Contradict} covers cases where the evidence is ambiguous, inconclusive, or does not distinguish the chosen answer from the correct one.

\emph{Rationale mode} asks whether the judge's rationale acknowledges the contradiction. The classifier sees the question, the chosen wrong answer, and the judge's rationale (but not the ground truth or the verifier Q\&A pairs), and labels the rationale as \textsc{Acknowledges} or \textsc{Does\_Not\_Acknowledge}. \textsc{Acknowledges} requires that the rationale (i)~contains concessive or adversative language (``although,'' ``however,'' ``despite,'' ``contradicts,'' ``while,'' ``but,'' ``some details differ,'' or similar), and (ii)~uses that language to engage with verification evidence opposing the chosen answer. The engagement may be vague (``although some details differ slightly''); what matters is that the rationale surfaces \emph{some} tension before dismissing it. Rationales whose only concessive language is unrelated to the verification evidence (e.g., generic hedging about uncertainty) do not qualify.

\paragraph{Classifier.} Both modes use Claude Opus 4.5 (\texttt{claude-opus-4-5-20251101}) at temperature 0 with a 500-token cap. The classifier sees the input described above with the system prompt for the relevant mode and is required to begin its response with one of two literal lines (\texttt{CLASSIFICATION: ACKNOWLEDGES} / \texttt{CLASSIFICATION: DOES\_NOT\_ACKNOWLEDGE} for rationale mode; \texttt{CLASSIFICATION: CONTRADICTS} / \texttt{CLASSIFICATION: DOES\_NOT\_CONTRADICT} for verification mode); cases where the response does not start with a recognized label are recorded as \textsc{Unclear}. Of the 90 cases, all received a clean classification in both modes. Each case is classified once; we did not run the classifier multiple times to estimate label variance.

\paragraph{Keyword baseline.} As a sanity check we also report a regular-expression baseline that flags any rationale containing one of the concessive tokens listed above. On the 90 wrong-answer cases the keyword baseline matches 50 (55.6\%); the LLM classifier matches 41 (45.6\%). The gap reflects formulaic hedging: rationales frequently contain concessive tokens used as stylistic templates without engaging with the verification evidence (e.g., ``although the question seems complex, the answer remains $D$''). The LLM classifier is required to verify the second condition (engagement with opposing evidence) and so produces a more conservative count. We report LLM-classifier numbers as the primary measurements throughout the paper.

\paragraph{Verification-mode result.} The verifier produces evidence that contradicts the chosen wrong answer in 89 of 90 cases (98.9\%); the single \textsc{Does\_Not\_Contradict} case is one where the judge's \textsc{No\_Mistake} verdict is in fact correct on the available evidence. This isolates the failure to the judgment stage: in 89 of 90 wrong-answer cases the contradicting evidence was present in the prompt and the nano judge nonetheless concluded \textsc{No\_Mistake}.

\paragraph{Cross-tabulation.} Table~\ref{tab:rationale-crosstab} reports the joint distribution of verification and rationale classifications across the 90 cases.

\begin{table}[!ht]
\centering
\small
\begin{tabular}{lcc}
\toprule
& \textbf{Evidence} & \textbf{Evidence does} \\
& \textbf{contradicts} & \textbf{not contradict} \\
\midrule
Judge acknowledges     & 41 & 0 \\
Judge does not ack.\   & 48 & 1 \\
\midrule
Total                  & 89 & 1 \\
\bottomrule
\end{tabular}
\caption{Cross-tabulation of verification-mode and rationale-mode classifications on the 90 wrong-answer cases from Run~3 of GPQA Diamond. The two failure-mode counts in Section~\ref{sec:gpqa-fiveconfig} (Finding~3) are read off the first column: 48 of 89 (53.9\%) \emph{Blindness}, 41 of 89 (46.1\%) \emph{Inability to act}.}
\label{tab:rationale-crosstab}
\end{table}

\section{GPQA Diamond Domain Breakdown (Cross-Model Configuration)}
\label{sec:appendix-gpqa-domains}

\begin{table}[!ht]
\centering
\small
\begin{tabular}{lrccc}
\toprule
\textbf{Domain} & \textbf{N} & \textbf{Init} & \textbf{Final} & \textbf{Net} \\
\midrule
Chemistry & 91 & 34.1\% & 59.3\% & $+23$ \\
Physics   & 71 & 76.1\% & 94.4\% & $+13$ \\
Biology   & 19 & 57.9\% & 42.1\% & $-3$  \\
\bottomrule
\end{tabular}
\caption{GPQA Diamond: per-domain results for the cross-model configuration (nano gen + gpt-5.2 verify/judge), aggregated to GPQA's three top-level domains. Chemistry combines organic and general chemistry; Physics combines general physics, quantum mechanics, astrophysics, and high-energy particle physics; Biology combines molecular biology and genetics. Per-subdomain breakdown is in the experiment report.}
\label{tab:gpqa-domains}
\end{table}

Chemistry and Physics are both strongly net positive, with Chemistry the largest absolute gain ($+23$). The Chemistry baseline (34.1\%) is the lowest of the three domains, and DISC closes more than half its gap to perfect accuracy, consistent with DISC being most effective where the initial model is weakest. Physics also gains substantially ($+13$) despite a much higher 76.1\% baseline, indicating the cross-model verifier finds errors even in the strongest-baseline domain. Biology is the exception: a small absolute net loss ($-3$), driven entirely by molecular biology ($n{=}15$, $-4$ net: 1 repair, 5 degradations), with genetics ($n{=}4$, $+1$ net) recovering one of those four. Molecular biology is the only subdomain where cross-model DISC under-performs the baseline; the small subdomain size makes the result sensitive to a few cases, but it is worth flagging as a domain where the failure pattern is qualitatively different from the broadly positive picture in chemistry and physics. A closer look at the five degradation cases is a natural follow-up.

\section{GPQA Judge Quality and Confusion Matrices}
\label{sec:appendix-confusion}

\begin{table*}[!ht]
\centering
\small
\begin{tabular}{lccccccccc}
\toprule
& \multicolumn{4}{c}{\textbf{Confusion matrix}} & \multicolumn{2}{c}{\textbf{Judge}} & \multicolumn{2}{c}{\textbf{Correction}} \\
\cmidrule(lr){2-5} \cmidrule(lr){6-7} \cmidrule(lr){8-9}
\textbf{Config} & \textbf{TP} & \textbf{FP} & \textbf{FN} & \textbf{TN} & \textbf{Prec.} & \textbf{Recall} & \textbf{TP fix \%} & \textbf{FP dest. \%} \\
\midrule
\textbf{Strong verify+judge} & \textbf{77} & \textbf{31} & \textbf{13} & \textbf{76} & \textbf{71.3\%} & \textbf{85.6\%} & \textbf{66.2\%} & \textbf{54.8\%} \\
All gpt-5.2                  & 31          & 19          & 18          & 130         & 62.0\%          & 63.3\%          & 58.1\%          & 52.6\% \\
\bottomrule
\end{tabular}
\caption{GPQA Diamond: judge classifier quality and downstream correction outcomes for the two configurations whose runs were re-executed with shared initial responses. Confusion-matrix counts are at the question level. Precision and recall are measured against ground truth (whether the current answer is wrong); TP fix\,\% is the rate at which true positives are correctly repaired; FP destruction\,\% is the rate at which false positives turn a correct answer incorrect. The all-nano and strong-verifier-only configurations are omitted because the nano judge says \textsc{No\_Mistake} on 100\% of examples, leaving precision, recall, and downstream rates undefined; the strong-judge-only configuration (Run~1, $+3$ net; Table~\ref{tab:gpqa-fiveconfig}) is omitted here because its confusion matrix was logged only in the prior non-shared-initial run and we report shared-initial data throughout this appendix.}
\label{tab:gpqa-judge-metrics}
\end{table*}

Comparing the cross-model row (gpt-5.2 judge on nano outputs) against the all-gpt-5.2 row (gpt-5.2 judge on its own outputs), every metric improves: precision rises from 62.0\% to 71.3\% ($+9.3$pp), recall from 63.3\% to 85.6\% ($+22.3$pp), TP fix\,\% from 58.1\% to 66.2\% ($+8.1$pp); FP destruction\,\% is essentially unchanged (52.6\% vs.\ 54.8\%). The largest single move is recall: with identical verifier and judge models, the gpt-5.2 judge flags an extra 22\% of errors when the answers it judges came from a different model. Precision also improves, so the additional flags are not just more noise. This is the unit-level signature of self-confirmation bias documented in Section~\ref{sec:gpqa-fiveconfig}: shared priors make a model less willing to flag errors in its own outputs, and the errors it does flag are less reliably error.

\section{Prompt Ablation: Full Results}
\label{sec:appendix-prompt-ablation}

Section~\ref{sec:prompt-ablation} reports the prompt-swap ablation: CoVe's single-pass gateless architecture with DISC's verification, evidence, and correction prompts. Table~\ref{tab:prompt-ablation} places it side-by-side with the three methods reported in Table~\ref{tab:main-results} on the same GPQA Diamond gpt-5.2 shared initial responses.

\begin{table}[!ht]
\centering
\small
\setlength{\tabcolsep}{4pt}
\begin{tabular}{lcccccc}
\toprule
\textbf{Method} & \textbf{Init} & \textbf{Final} & \textbf{Imp} & \textbf{Deg} & \textbf{I:D} & \textbf{Net} \\
\midrule
DISC                                & 75.3 & \textbf{79.3} & 18 & 10  & \textbf{1.80} & \textbf{$+8$}  \\
Self-Refine                         & 75.3 & 75.3          & 13 & 13  & 1.00          & $\phantom{+}0$  \\
CoVe                                & 75.3 & 69.7          & 17 & 28  & 0.61          & $-11$ \\
CoVe arch + DISC prompts            & 75.3 & 67.2          & 19 & 35  & 0.54          & $-16$ \\
\bottomrule
\end{tabular}
\caption{Prompt-swap ablation on GPQA Diamond with gpt-5.2 ($N{=}198$). The last row is the ablation: CoVe's single-pass gateless architecture with DISC's verification, evidence, and correction prompts. The first three rows reproduce the GPQA Diamond gpt-5.2 cells of Table~\ref{tab:main-results} for side-by-side comparison. Init, Final are accuracy (\%); Imp, Deg are example-level counts; I:D is the precision-style ratio; Net = Imp $-$ Deg.}
\label{tab:prompt-ablation}
\end{table}

Repair rate (the recall-style metric $I/N_{\text{wrong init}}$) for the prompt-swap ablation is 38.8\%, comparable to DISC's 36.7\%, indicating that DISC's prompts continue to surface and fix errors even inside CoVe's gateless architecture. Preservation rate (fraction of initially-correct answers kept correct) drops to 76.5\% vs.\ DISC's 93.3\%; without the gate, the committed-corrector framing rewrites already-correct answers as if they contained a confirmed error.

\section{Licenses and Terms of Use}
\label{sec:appendix-licenses}

All datasets used in this work are publicly released for research use and do not contain personally identifying information about private individuals or offensive content. Our use is consistent with their licenses: BIG-Bench Mistake
(Apache 2.0), HotpotQA (CC BY-SA 4.0), GPQA Diamond (MIT). 
We reimplemented CoVe and Self-Refine from their paper descriptions.
The models used as study subjects (GPT-4o, gpt-5.2, gpt-4.1-nano,
Claude Sonnet 4.5) were accessed under the OpenAI and Anthropic API
terms of service.

External libraries: \texttt{datasets} (v$\geq$4.4.1) and \texttt{numpy} (v$\geq$1.24.0). 

\end{document}